# TGTV: An Interpretable Framework for UAV Fault Detection and Traceability via Spatiotemporal Graph Learning and Large Language Models

Zixuan Wang, Changbo Shi, and Ji Liu[1]

Abstract—Fixed-wing unmanned aerial vehicles experience operating-condition transitions, asynchronous sensor sampling, and control-system compensation that can obscure fault signatures. Existing temporal detection methods have limited capacity to represent coupled faults across subsystems, track their evolution, and constrain language-model diagnoses with quantitative evidence. This paper proposes TGTV, an interpretable fault detection and traceability framework that integrates spatiotemporal graph learning with a large language model. A temporal convolutional network and a graph attention network with weighted fused adjacency extract multiscale temporal patterns and spatial dependencies among sensors. A hybrid temporal Transformer and variational autoencoder then evaluates window-level anomalies using prediction errors and reconstruction probabilities. Continuous-time Shapley attribution identifies critical sensors and high-contribution intervals while preserving the temporal continuity of fault evolution. A structured evidence package combines detection results, graph relationships, temporal attribution, sensor attribution, and operating context. An evidence-constrained diagnosis protocol uses this package to generate flight-level conclusions with traceable evidence identifiers. Cross-flight holdout experiments on the public ALFA fixed-wing flight dataset yield a window-level area under the precision–recall curve of 0.976, accuracy of 0.973, recall of 0.902, and F1 score of 0.920. Flight-level fault diagnosis achieves recall of 0.943 and an F1 score of 0.944. These results demonstrate accurate anomaly detection and quantitatively supported fault-tracing cues, providing a basis for interpretable post-flight fault review and analysis.

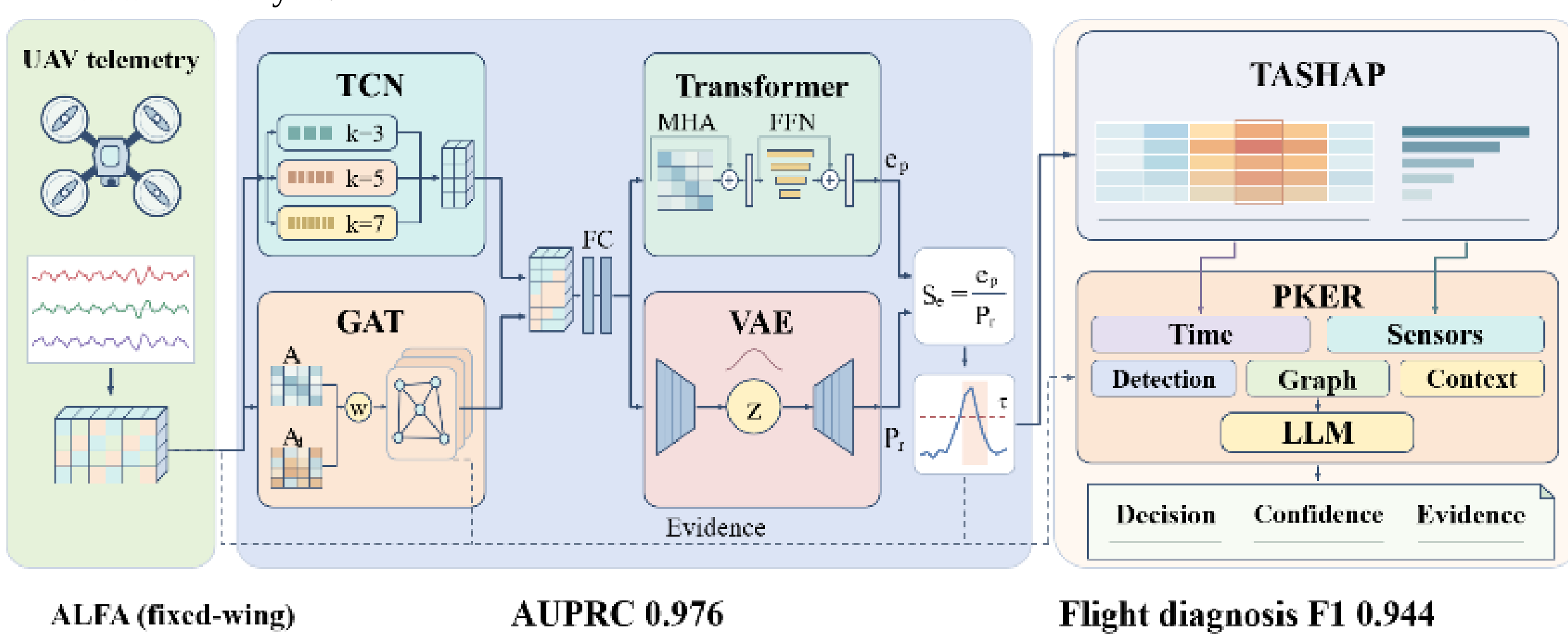

## I. Introduction

WITH the rapid development of the low-altitude economy and the iterative upgrading of unmanned aerial vehicle (UAV) technology, UAVs, featuring long endurance range, wide flight coverage, strong load capacity and high

[1]Corresponding author: Ji Liu (e-mail: ji.l@sydney.edu.au).

Zixuan Wang and Changbo Shi are with the School of Computer Science, Faculty of Engineering, The University of Sydney, Sydney, NSW 2006, Australia.

Ji Liu is with the School of Electrical and Computer Engineering, Faculty of Engineering, The University of Sydney, Sydney, NSW 2006, Australia.

operational efficiency, have been widely deployed in key civil and industrial scenarios including power line inspection, topographic and geomorphological mapping, geological hazard investigation, emergency rescue reconnaissance, agricultural and forestry plant protection, and meteorological monitoring. They have become core equipment for intelligent low-altitude operations [1]. During practical mission flights, the propulsion subsystem, actuation subsystem, navigation and control subsystem, and environmental perception subsystem of a UAV constitute a highly coupled closed-loop control system. Tight state correlations and dynamic feedback mechanisms exist among these subsystems and sensors. Flight states are highly susceptible to the combined effects of multiple factors such as complex airspace environments, dynamic mission commands, equipment aging and wear, and external airflow disturbances [2]. Compared with rotary-wing UAVs, fixed-wing UAVs fly at higher speeds, undergo more frequent operating condition transitions, and impose stricter requirements on the precision of the control closed loop. Their fault propagation exhibits stronger concealment, coupling and hysteresis. If minor local equipment anomalies cannot be identified in a timely manner, they will be continuously amplified through active compensation and regulation of the control system, and gradually spread across the entire system. This may trigger severe issues such as flight attitude deviation, trajectory drift, power instability and control failure, ultimately resulting in mission failure or even crash accidents, which cause multiple losses including equipment damage, data loss and potential airspace safety hazards. Therefore, high-precision, interpretable and traceable flight anomaly and fault detection technologies serve as the core enablers to guarantee safe and stable UAV operation, improve mission execution reliability and reduce operation and maintenance costs [3].

In the field of UAV engineering operation and maintenance, conventional fault monitoring and diagnosis paradigms mostly rely on post-hoc manual analysis, fixed threshold discrimination and anomaly screening using single parameters, which suffer from notable technical limitations. Post-hoc manual analysis is heavily dependent on the engineering experience of maintenance personnel, featuring strong subjectivity, low efficiency and an inability to support real-time monitoring. With growing demands for intelligent UAV operation and maintenance, time-series anomaly detection methods based on deep learning have gradually become the mainstream of research. Such methods can effectively mine latent anomalous features within flight telemetry data and realize automated and intelligent fault identification. Nevertheless, existing technical solutions still confront three core bottlenecks during deployment in real engineering scenarios, which severely restrict the practicality and reliability of UAV fault detection and traceability [4]. First, the capability to characterize cross-subsystem coupled faults under complex and dynamic operating conditions is insufficient. UAV flight operating conditions cover multiple typical states including takeoff, climb, cruise, turn, dive and landing. The control logic, sensor operating status and subsystem coupling relationships vary significantly across different working conditions. External airflow disturbances and changes in mission commands can further reconstruct the correlation and dependency relationships among variables. Most existing studies perform anomaly modeling under steady-state conditions and fixed topological relationships, treating sensor parameters as independent time-series variables. Consequently, these methods struggle to accurately capture coupled anomalous features across sensors and subsystems, leading to high false alarm and missed detection rates. The generalization ability of such models fails to adapt to real flight scenarios with multiple operating conditions [5]. Second, there is a lack of modeling for the temporal evolution of faults and fine-grained attribution. The vast majority of UAV flight faults do not occur instantaneously. Existing time-series anomaly detection and attribution methods mostly compute features and analyze contribution weights based on isolated sampling points or fixed static windows. This approach breaks the temporal continuity of fault evolution and cannot fully capture the contextual information throughout the entire fault progression from latent to manifest states. It is difficult to precisely locate the initial fault onset time, key evolutionary nodes and compensation failure instant. As a result, fault attribution outputs are fragmented and lack fine granularity, which cannot satisfy the engineering requirements for fault traceability and analysis [6]. Third, fault detection results suffer from weak interpretability, and traceability diagnosis lacks quantitative evidence constraints. Some recent studies have adopted large language models (LLMs) to generate natural-language explanations for anomaly outputs, which effectively improves the readability of model predictions. However, these approaches generally exhibit the limitation of “prioritizing explanation over evidence”. Explanations generated by LLMs mostly rely on the intrinsic semantic generation capability of the model, without rigid constraints from real flight numerical data, structured correlation relationships and quantitative metrics. This greatly limits the trustworthiness and practicality of intelligent detection models for industrial deployment [7].

To address the aforementioned issues, the main contributions of this paper are as follows:

(1)A spatiotemporal feature extraction method based on TCN-GAT is proposed. To address the challenges of strong temporal dependence in multivariate UAV telemetry data, complex spatial coupling among sensors, and the limitation that a single model cannot fully mine hidden information, multi-scale TCN and multi-head GAT are introduced to capture the temporal and spatial features of flight data, respectively. This enables deep mining of data information and provides inputs for anomaly detection discrimination.

(2) A hybrid anomaly detection model based on Transformer and VAE is constructed. To tackle the poor performance of conventional prediction-based methods for long sequences, Transformer and Variational Autoencoder (VAE) are integrated. Leveraging the self-attention mechanism to model long-range temporal dependencies, the proposed method retains the merits of sequence modeling while maintaining high sensitivity for anomaly identification. Anomaly indicators are established from prediction errors and reconstruction probabilities to precisely adapt to dynamically correlated scenarios under multiple working conditions.

(3)A continuous-time Shapley attribution mechanism for the full lifecycle of faults is proposed. To address the drawback that traditional point-wise attribution cannot adapt to the continuous evolution characteristics of faults, the calculation paradigm of isolated sampling point contributions is abandoned. Based on complete fault detection windows, an accurate correspondence between the temporal evolution of faults and model anomaly judgment is established to achieve fine-grained temporal attribution.

(4)A voyage-level traceability diagnosis method with multi-dimensional evidence constraints based on LLM is designed. To tackle the problems that existing LLM-based anomaly explanations lack quantitative constraints and traceability, a five-dimensional structured evidence system is constructed, including D1 fault discrimination evidence, G1 graph-edge correlation evidence, T1 time-period evidence, S1 sensor anomaly evidence, and C1 working-condition constraint evidence. Standardized interpretable output rules and an evidence ID reference mechanism are formulated to generate complete voyage-level fault diagnosis records, enabling evidence-based and traceable LLM explanations.

The research value of this paper lies in integrating detection and attribution into a continuous methodological pipeline, demonstrating the complete procedure from anomaly discovery to fault analysis. The remainder of this paper is organized as follows. Section II reviews related work. Section III elaborates on the proposed methodology. Section IV presents the analysis of experimental results for anomaly detection. The final two sections discuss the experimental findings and draw conclusions.

## II. Related works

UAV fault detection and traceability is an interdisciplinary research area integrating UAV health management, multivariate time-series anomaly mining, spatiotemporal graph structure modeling, interpretable machine learning, and knowledge enhancement of large language models (LLMs). Existing relevant studies have been gradually carried out in the directions of deep anomaly detection for multivariate time series, temporal attribution, and LLM-aided diagnosis with knowledge enhancement, laying a relatively solid technical foundation. This paper systematically reviews the state-of-the-art research at home and abroad, summarizes the strengths and limitations of existing techniques, and clarifies the differentiated positioning and core breakthroughs of this research.

UAV flight telemetry data are essentially high-dimensional, nonlinear, strongly correlated and continuously time-series multivariate data, consisting of dozens of core variables such as attitude angles, angular velocities, flight speeds, control surface deflections, engine rotational speeds, and sensor measurements. Multivariate time-series anomaly detection serves as the core technical support for UAV fault identification [7]. In recent years, with the rapid advancement of deep learning, multivariate time-series anomaly detection has undergone iterative upgrades from traditional statistical modeling and shallow machine learning to deep representation learning. Multiple mature technical paradigms have been developed, including probabilistic modeling, reconstruction learning, attention-based representation, and multi-scale correlation mining. These approaches can effectively capture latent anomalous features within high-dimensional time-series data and have been widely adopted in industrial equipment monitoring, rail transit maintenance, and aerospace condition perception [8].

To address the limitation that traditional sensor-data-and-neural-network-based UAV anomaly detection methods struggle to fully extract fault features under complex environments, Yang et al. proposed a multimodal neural network model FMDNN based on heterogeneous sensor data. The model uses MLP and CNN to extract nonlinear features and critical structural features in parallel. After feature fusion, a fully connected layer and Softmax classifier are adopted to judge UAV health status, and its favorable robustness is verified on classification tasks with noisy data [9]. He et al. adopted a flight-level validation scheme to avoid data leakage and proposed a grouped deformable Transformer model FG-DeformT to capture heterogeneous local temporal shifts. Experiments on the ALFA dataset demonstrate the robustness of this model [10]. To tackle the scarcity of anomalous flight samples and the difficulty in extracting spatiotemporal coupling features of flight parameters in UAV anomaly detection, Feng et al. put forward an unsupervised causality-enhanced graph neural network (CEG). It leverages causality-enhanced graph attention to extract spatial features and trend decomposition to capture temporal dependencies. Low-rank regularization and residual adaptive bidirectional smoothing are used for noise reduction, enabling accurate identification of various UAV faults including engine, control surface and aileron faults [11]. To improve UAV safety, Hu et al. presented the FastICA-TGAK-OCELM anomaly detection method. FastICA is employed to reconstruct UAV data, and a triangular global alignment kernel replaces the kernel of the one-class extreme learning machine to effectively detect point anomalies [12]. Faced with the challenges of sample scarcity, inconsistent sampling frequencies, missing sensor readings and temporal non-stationarity in UAV fault diagnosis, Liu et al. proposed an unstable Transformer framework UTIDH with imbalanced data processing. Integrating multi-task Gaussian process imputation and unstable attention mechanism, the framework achieves remarkable accuracy improvement compared with baseline models [13]. Chen et al. combined accelerated fuzzy entropy, self-loop decaying masked quadratic graph attention mechanism and GRU, and proposed the UAV-EGAT model. This method achieves a fault detection accuracy of 97.3% for elevator faults and 82.6% on the ALFA dataset, supporting real-time UAV fault detection [14]. Nevertheless, the detection performance of deep learning models is highly dependent on training supervision strategies, temporal window definitions, anomaly threshold settings and data partitioning schemes. Different parameter configurations and data splitting strategies can lead to substantial performance fluctuations, making it difficult to guarantee the generalization and stability of algorithms [15].

There exist inherent topological correlations and dynamic dependencies among various sensors and subsystems of UAVs [16]. Pure temporal modeling can only mine the temporal features of data and fails to characterize spatial correlations between variables as well as fault propagation structures. Spatiotemporal graph modeling and causal structure learning techniques provide core solutions for such correlation characterization problems and have been widely applied in anomaly detection and fault propagation analysis of complex industrial systems [17]. To address the limitation that existing unsupervised multivariate time-series anomaly detection methods struggle to model dynamically evolving spatiotemporal dependencies and effectively eliminate confounding interferences, Chen et al. proposed a spatiotemporal causal inference detector (SCID). It achieves fine-grained spatiotemporal dependency modeling via counterfactual causal decoupled encoding, dual-mask predictive reconstruction, and dual-granularity optimization constraints, and obtains state-of-the-art detection performance by comparing with twelve baseline methods across five datasets [18]. Jiang et al. presented an integrated causal anomaly detection framework (ICAD). A Granger structural causal model is used to construct spatiotemporal causal representations and a causal anomaly scoring mechanism. Graph pruning and parallel optimization are adopted to improve computational efficiency. The framework achieves a lower false alarm rate and stronger noise robustness, and supports high-precision fault root cause localization, which effectively addresses the challenges of difficult anomaly detection, insufficient interpretability and scalability in distributed systems [19]. Chen et al. developed a counterfactual causal reinforcement learning anomaly detector (Tri-CRLAD). By integrating causal feature extraction and a triple decision support mechanism, the detector exhibits stronger robustness and smaller performance degradation under scenarios with scarce labeled anomaly samples, improving model flexibility [20]. Xing et al. designed a dual-channel deep learning framework combining temporal convolutional networks and variational autoencoders. Contrastive learning is adopted to construct a unified indicator representation, and causal reasoning is used to trace fault propagation paths. This framework can greatly shorten fault troubleshooting time, reduce mislocalization rate, and satisfy the computing resource requirements for real-time monitoring [21]. Ali et al. integrated dynamic global-local ordinal-preserving projection, adaptive neuro-fuzzy inference system and causal lineage graphs to build a hybrid causal neuro-fuzzy framework. It realizes robust anomaly detection, accurate fault isolation, root cause localization and key variable identification, which is suitable for complex industrial monitoring scenarios [22]. Zeng et al. proposed a causal grey wolf optimization framework (CGWO). Real causal features are screened through causal weight transformation, potential outcome modeling and feature optimization balancing accuracy dimensions. The framework possesses high interpretability, computational efficiency and scalability for edge deployment [23]. Although structure learning techniques enable interpretable modeling of system correlations, they suffer from obvious adaptation limitations in real UAV flight scenarios.

With the widespread adoption of deep learning models in anomaly detection, the issues of unexplainable predictions and untrustworthy decisions arising from the "black-box" nature of models have become increasingly prominent. Interpretable temporal attribution techniques and knowledge-enhanced large model approaches have emerged as active research frontiers in the field of intelligent fault detection [24]. Temporal attribution techniques aim to identify the core rationale behind model decisions by quantifying the contributions of individual input variables and temporal segments to the model's anomaly judgment results. Zhao et al. proposed IMDM-PC, a physically constrained multi-scale large language model. Integrating incremental multi-scale decomposition hybrid encoding, physically consistent residual correction, and an adaptive threshold strategy for online residual envelopes, this model achieves streaming anomaly detection and effectively reduces false alarm and missing alarm rates [25]. Jiang et al. developed GCRdetector, a model based on graph contrastive representation learning. By leveraging intra-view learning and a dual-view contrastive alignment mechanism to strengthen instance self-consistency, the model achieves competitive performance and strong anti-interference capability on both real-world and simulated datasets. It can effectively support large language model enabled safety monitoring tasks and improve interpretability [26]. Zhang et al. designed the LLM-LADE framework built upon large language models. A three-stage collaborative architecture consisting of data augmentation, parameter-efficient fine-tuning, and online optimization is constructed to realize online adaptation without parameter updates. The proposed method outperforms existing algorithms in detection performance, and the generated explanations are informative and human-readable [27]. Lau et al. presented BiTAD, an anomaly detector based on bidirectional temporal attention modeling, and devised the TwinLens anomaly inference paradigm that integrates SHAP and TimeSHAP. Using complete time-series sequences as the basis, this paradigm quantifies the contributions of different temporal regions to model outputs and implements global temporal attribution [28]. Anees et al. established a wind power prediction framework that integrates random forest prediction, SHAP-based interpretability analysis, and safety verification. Experiments demonstrate that the framework achieves favorable prediction accuracy, interpretable feature attribution, satisfactory adversarial robustness, and competent anomaly detection capability, and can shorten the time cost of fault diagnosis [29]. Nevertheless, existing studies directly feed model outputs into LLMs for open-ended reasoning, without external quantitative evidence, structural constraints, or traceable citation mechanisms. The explanations generated by LLMs rely on subjective semantic generation of the model, which suffers from arbitrary reasoning, large deviations from ground-truth data, and the lack of quantitative verification, making them incapable of meeting the requirements of standardized review in engineering scenarios.

## III. Research methodology

As illustrated in Fig. 1, the proposed method consists of three main stages: data preprocessing and spatiotemporal feature extraction, Transformer-VAE hybrid anomaly detection, continuous temporal attribution, and evidence-based diagnosis.

1. Spatiotemporal feature extraction. Multi-source telemetry data are cleaned independently for each voyage. On this basis, TCN-GAT, composed of multi-scale temporal convolutional networks and multi-head graph attention networks, is adopted to extract temporal dependencies and cross-sensor spatial coupling features within the window. The resulting spatiotemporal features characterizing flight states serve as inputs for anomaly discrimination and relationship tracing.

2. Transformer-VAE hybrid anomaly detection. The spatiotemporal feature matrix is fed into the temporal Transformer prediction network and the VAE reconstruction network to calculate prediction errors and reconstruction probabilities, respectively. A window-level anomaly indicator is constructed accordingly. This indicator is used for threshold judgment in the main detection pipeline, and the detection outputs together with subsequent explanatory evidence are recorded separately.

3. Continuous temporal attribution and evidence-based diagnosis. TASHAP is used to compute continuous Shapley contributions by taking complete fault evolution periods as computation units, so as to locate high-contribution periods and critical sensors. Furthermore, PKER organizes detection evidence, graph relation evidence, time-period evidence, sensor evidence and working-condition evidence into a five-dimensional structured evidence package. It constrains the large language model to output fault judgments and tracing reasoning with evidence identifiers and data support.

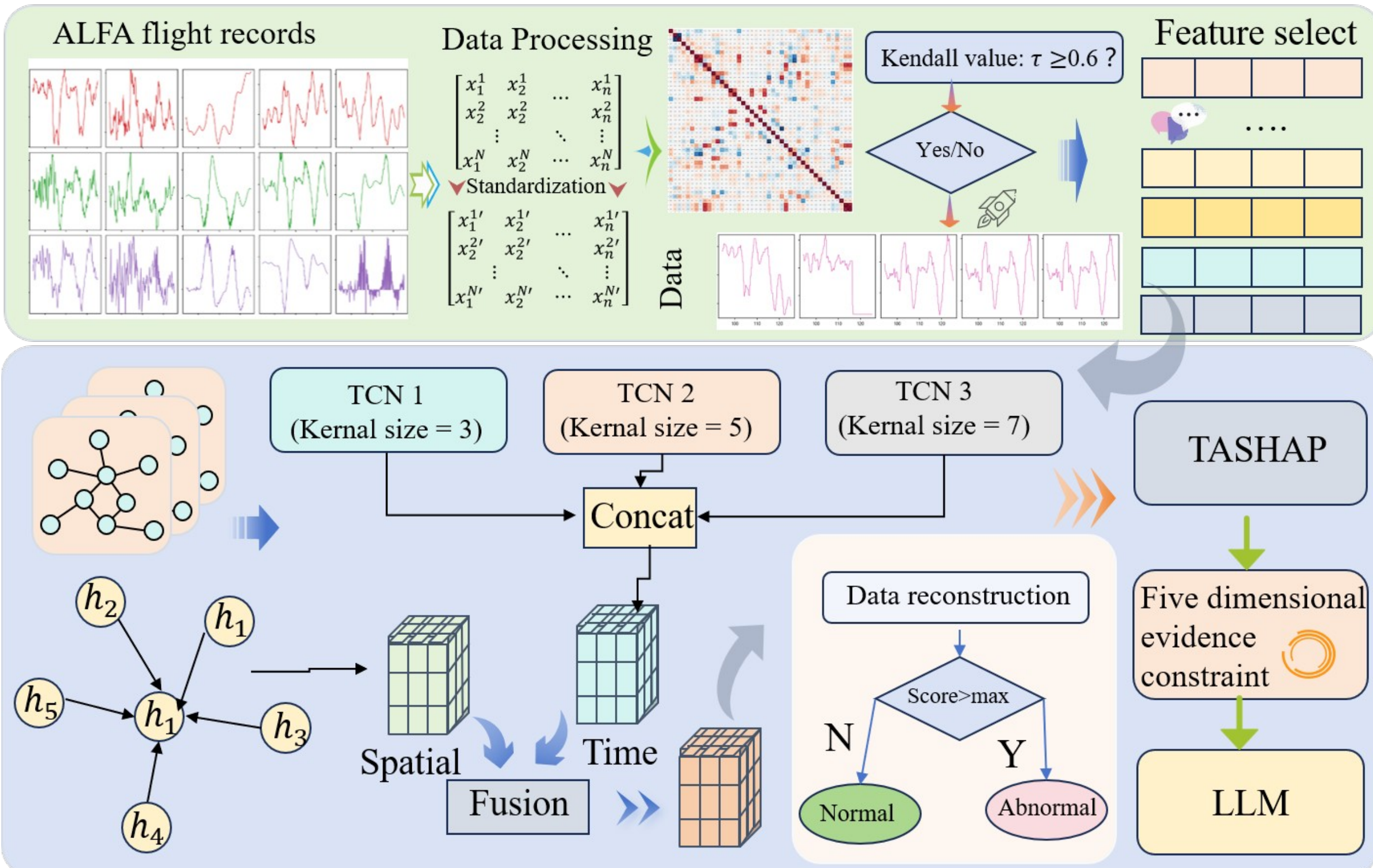

Fig. 1. Overview of the proposed UAV fault detection and traceability framework.

The detailed pseudocode of the algorithm is presented in Algorithm 1.

| **Algorithm1** | **Overall Detection and Diagnosis Pipeline** |
|---|---|
| Input | Voyage-partitioned datasets $D_{tr}$, $D_{val}$, $D_{te}$, candidate window selection rules |
| Output | Window anomaly score $S_e$ and label $\hat{y}$, time segment and sensor attribution, voyage diagnosis results |
| 1 | Independently clean, align telemetry data and interpolate to fill missing observations per voyage; |
| 2 | Fit normalization parameters using $D_{tr}$, then apply these fixed parameters to $D_{val}$ and $D_{te}$; |
| 3 | Construct sliding windows with window length L=40 and step size 5, and extract spatio-temporal features $\boldsymbol{H}$; |
| 4 | Train the Transformer-VAE model, and determine threshold τ on validation voyages; |
| 5 | **FOR** each window $X_w$ in test voyages **DO** |
| 6 | Compute $\boldsymbol{H}$, graph adjacency $\boldsymbol{A}$, anomaly score $S_e$ and prediction label $\hat{y}$; |
| 7 | **END FOR** |
| 8 | **FOR** candidate windows $X_w$ selected by predefined rules **DO** |
| 9 | Read graph module output $G_1$, obtain $T_1$ and $S_1$; |
| 10 | Form an evidence package consisting of detection summary $D_1$, operating condition summary $C_1$, together with $G_1$, $T_1$, $S_1$; |
| 11 | Generate diagnosis status and reasoning with evidence ID, and associate them with the corresponding voyage; |
| 12 | **END FOR** |
| 13 | **RETURN** window detection results, continuous time-segment attribution and voyage diagnosis results. |

### *A. TCN-GAT Spatiotemporal Feature Extraction*

UAV flight telemetry data are high-dimensional, and issues such as network interruptions, sensor asynchronization, and data loss occur during data acquisition. Therefore, feature extraction is required for the collected data to improve the accuracy of anomaly detection. The workflow is illustrated in the Fig. 2.

To fully capture multi-scale temporal dependencies, this paper designs multiple TCNs with different receptive field sizes to extract temporal features from the telemetry data. Each TCN contains three temporal convolution modules; each module is composed of two dilated causal convolutions with weight normalization, and the output is activated by the ReLU function. The dilated causal convolution of the TCN is formulated as

$$F(t)=\sum_{i=0}^{k-1} f(i)\, x_{t-di} \tag{1}$$

where $F(t)$ denotes the convolution result of the element at time $t$; $f(i)$ is the convolution kernel; $k$ is the kernel size; and $d$ is the dilation factor. The right-end cropping of the causal convolution ensures that the output at time $t$ depends only on the sampling points at time $t$ and before, avoiding future information leakage. The TCN transmits the result of the upper layer to the lower layer through residual connections:

$$y_n=F(t)+\sum_{i=1}^{n-1} R(t_i, F_i) \tag{2}$$

where $y_n$ is the temporal feature extracted by each TCN layer; $F$ denotes the dilated convolution operation; and $R$ denotes the cross-layer residual connection. The residual connection passes the upper-layer features to the lower layers, alleviating the gradient degradation of deep networks. The output of each TCN is concatenated by the concat(·) function, and is then fed into a pooling layer and a fully connected layer to obtain the temporal feature matrix:

$$H_t=FC\left(Avgpooling\left(concat\left(h_t^1,\dots,h_t^c\right)\right)\right) \tag{3}$$

where $h_t^1,\dots,h_t^c$ are the temporal features output by the $c$ TCN branches; concat(·) denotes feature concatenation; Avgpooling(·) denotes average pooling; FC(·) denotes the fully connected layer; and $H_t$ is the temporal feature matrix after multi-scale fusion.

Define $H=[h_1,h_2,\dots,h_n]$, with $h_n\in\mathbb{R}^D$, as the set of nodes, where $n$ is the number of sensor data types and $D$ is the feature dimension of the node vectors. The attention scores are computed by the following formulas:

$$\begin{cases} e_{ij}=LeakyReLU\left(\left[a^T W h_i \,\|\, W h_j\right]\right) \\ \alpha_{ij}=\dfrac{\exp(e_{ij})}{\sum_{k\in C_i}\exp(e_{ik})} \end{cases} \tag{4}$$

where $W\in\mathbb{R}^{D'\times D}$ is the weight matrix and $D'$ is the feature dimension of the output vectors; $a^T$ is the column vector of learnable parameters; LeakyReLU is the activation function; and $e_{ij}$ is the attention relevance between nodes $i$ and $j$, which is normalized by Softmax to obtain the attention score $\alpha_{ij}$.

To achieve favorable feature extraction, this paper combines the multi-head attention mechanism and adds multiple parallel attention score computation processes, and finally obtains the feature matrix of node $i$:

$$h_i=\sigma\left(\frac{1}{K}\sum_{k=1}^{K}\sum_{j\in C_i}\alpha_{ij}^k W^k h_j\right) \tag{5}$$

where $K$ is the number of attention heads; $\alpha_{ij}^k$ and $W^k$ are the attention score and the weight matrix of the $k$-th attention head, respectively. The multi-head mechanism characterizes the node relationships in parallel from multiple representation subspaces, and the fused result yields a more robust spatial feature matrix $H_s = [h_1, h_2, \ldots, h_i]$.

The attention scores are computed dynamically with the input features, and the dependencies among nodes adaptively change with the flight conditions, forming a conditional dynamic graph. Meanwhile, to fuse prior structural knowledge, a static topology prior is further introduced: each sensor $j$ corresponds to a learnable node embedding $e_j \in \mathbb{R}^h$; after all node embeddings constitute $E$, the static adjacency is defined as $A_s = softmax\left(E E^T / \sqrt{h}\right)$, and the conditional dynamic adjacency $A_d(X)$ is generated by a linear projection of the node statistics within the window; the static and dynamic adjacencies are fused in a weighted manner:

$$A_d(X) = softmax\left(\frac{Q(N(X))K(N(X))^T}{\sqrt{h}}\right), A = (1 - w_d) A_s + w_d A_d \tag{6}$$

where $E$ is the matrix composed of all learnable node embeddings and $h$ is the node embedding dimension; $A_s$ is the static adjacency matrix obtained from the inner products of nodes and shared across windows; $N(X)$ denotes the statistics of each node within the window, including the mean, standard deviation, final value, and the difference between the first and final values; $Q(\cdot)$ and $K(\cdot)$ are the linear projections of the node statistics; $A_d(X)$ is the conditional dynamic adjacency matrix that varies with the window; $w_d = \sigma(a_d)$ is the learnable dynamic graph weight; and $A$ is the weighted fusion of the static and dynamic adjacencies. The attention edge weights of the fused adjacency change with the flight phase and local anomalies, constituting interpretable graph-relationship evidence.

The obtained temporal and spatial features are fed into fully connected layers for fusion and dimensionality reduction, and the spatio-temporal feature matrix $H$ that characterizes the flight state of the UAV is obtained:

$$H = FC2\left(FC1\left(H_t, H_s\right)\right) \tag{7}$$

where $H_t$ and $H_s$ are the temporal and spatial feature matrices extracted by the TCN and the GAT, respectively; FC1 and FC2 are fully connected layers that accomplish feature fusion and dimensionality reduction; and $H$ is the spatio-temporal feature matrix characterizing the flight state. The pseudocode of the specific algorithm is shown in Algorithm 2.

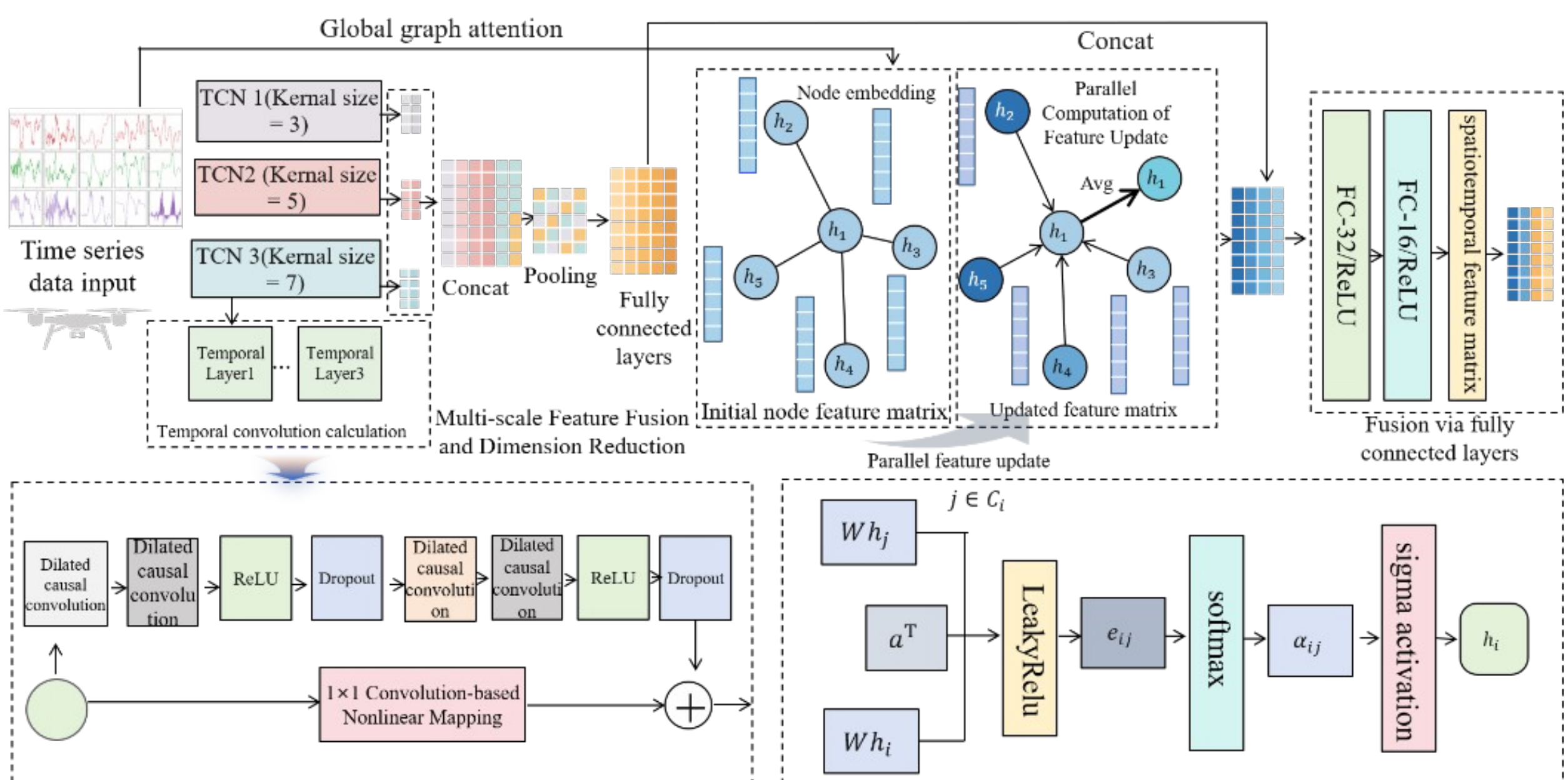

Fig. 2. Spatiotemporal feature extraction for UAV telemetry.

**Algorithm2** **TCN-GAT Spatio-Temporal Feature Extraction**

Input: Voyage telemetry sequence X, normalization parameters from training set, TCN-GAT parameters and node embedding E

Output: Window set $\{X_w\}$, spatio-temporal features $\{\boldsymbol{H}_w\}$ and graph relation $\{\boldsymbol{A}_w\}$

```
FOR each window X_w DO
    FOR each TCN branch DO
        Perform dilated causal convolution, activation and residual connection to obtain temporal features of the branch;
    END FOR
    Concatenate the outputs of branches, and obtain H_t via pooling and fully-connected mapping;
    Calculate A_s from E, and calculate A_d(X_w) from node statistics N(X_w);
    Calculate A_w=(1−w_d)A_s+w_d A_d(X_w) according to Eq.(6);
    Compute multi-head graph attention according to Eqs.(4)–(5) to obtain H_s;
    Fuse H_t and H_s according to Eq.(7) to obtain H_w;
END FOR
RETURN {X_w}, {H_w}, {A_w}.
```

### *B. Anomaly Detection Based on Hybrid Neural Network*

In this paper, a hybrid anomaly detection model is constructed by combining the temporal Transformer and variational autoencoder (VAE). Taking the spatio-temporal feature matrix $H$ as input, the model obtains prediction errors and reconstruction probabilities, based on which an anomaly indicator is established. The framework is illustrated in the Fig. 3.

The prediction network adopts a temporal Transformer, which exploits the self-attention mechanism to capture the long-range dependencies of the spatio-temporal feature sequences, so as to map the nonlinear relationship between the spatio-temporal feature data and the flight state data. Suppose the input sequence is projected through learnable linear projections to obtain the query matrix $Q$, the key matrix $K$, and the value matrix $V$; after the weighted summation of self-attention, the prediction result is output. The computation process is formulated as

$$y_p = f\left(softmax\left(\frac{QK^T}{\sqrt{d_k}}\right)V\right) \tag{8}$$

where $Q$, $K$, and $V$ are the query, key, and value matrices obtained by the learnable linear projections of the input sequence, respectively; $d_k$ is the dimension of the key vectors, which is used to scale the attention weights; and $f$ denotes the feed-forward network and the activation function. The Transformer captures temporal dependencies in parallel from multiple representation subspaces through multi-head self-attention; residual connections and layer normalization alleviate the gradient degradation of deep networks, and Dropout is employed to suppress overfitting. After the model is trained, feeding the test set into the model yields the predicted values. To realize anomaly detection, this paper takes the residual $e_p$ between the predicted values and the actual values as the output of the prediction network, i.e.,

$$e_p = \frac{1}{M}\sum_{i=1}^{M}\left\|y_i - y_{p,i}\right\|^2 \tag{9}$$

where $M$ is the number of prediction samples; $y$ denotes the actual observed values in the window; $y_p$ denotes the predicted values; and $e_p$ is the prediction residual.

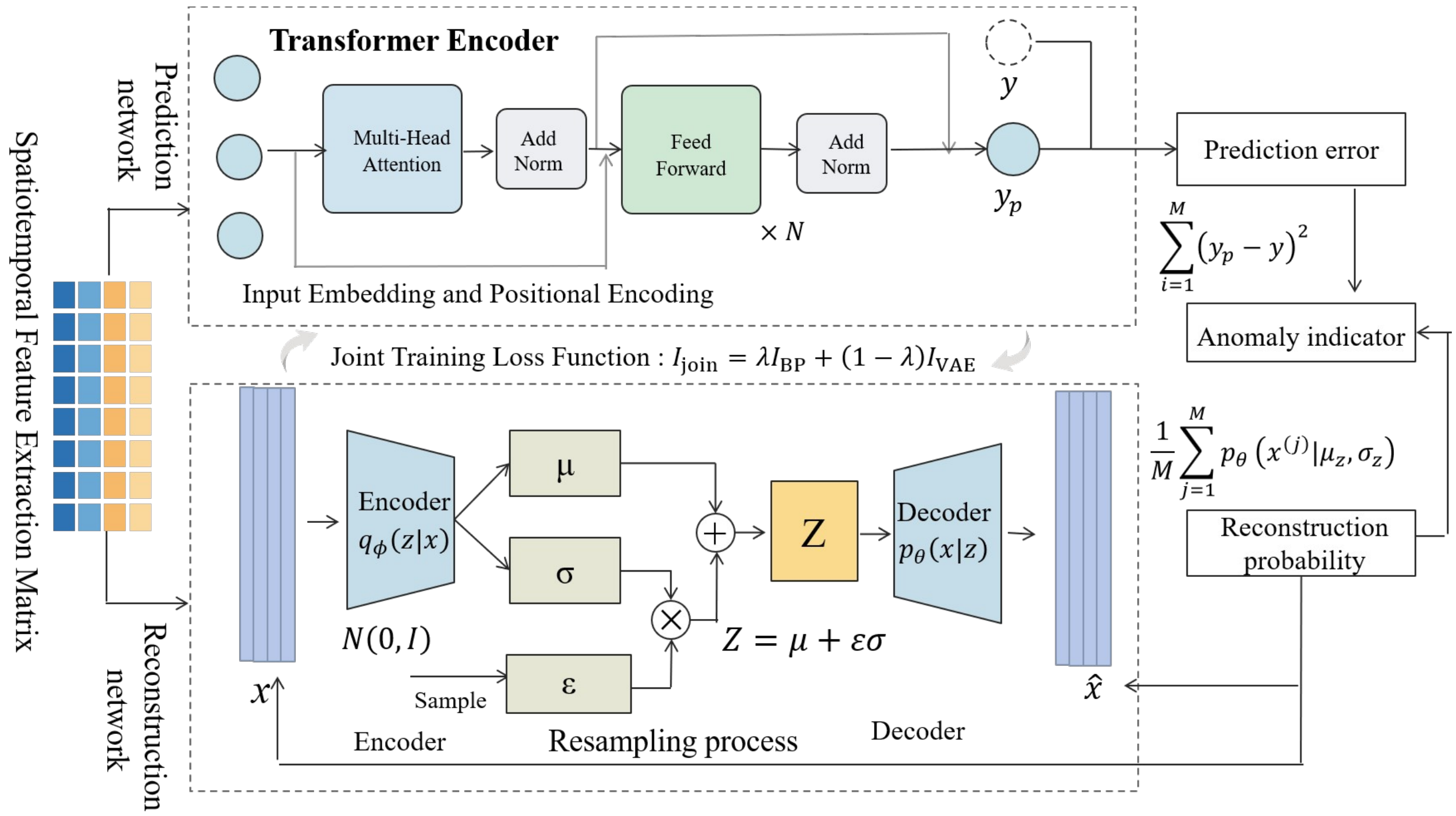

Fig. 3. Flowchart of anomaly detection for the hybrid neural network.

The reconstruction network adopts a VAE as the reconstruction network of the detection model, which mainly consists of an encoder and a decoder. In this paper, the encoder and the decoder of the VAE are structurally symmetric neural networks. Let the input data be $x$; the latent variable $z$ is obtained through the encoder with parameters $\phi$, which can be expressed as the distribution $q_\phi(z|x)$. The computation of the latent variable $z$ is formulated as

$$z = \mu + \varepsilon\sigma \tag{10}$$

where $\mu$ and $\sigma$ are the mean and the standard deviation solved by the encoder, respectively; and $\varepsilon$ is the noise introduced according to the standard normal distribution. Subsequently, the decoder receives the latent variable $z$ and outputs the reconstructed data $\hat{x}$ with parameters $\theta$, expressed as the distribution $p_\theta(\hat{x}|z)$.

When the input data deviate from the normal pattern, the VAE is usually unable to reproduce the original data. This paper measures the reconstruction result of the data through the reconstruction probability, and distinguishes normal data from anomalies by outputting the reconstruction probability of each data sample. The reconstruction probability is computed as

$$P_r = \frac{1}{M} \sum_{j=1}^{M} p_\theta \left( x^{(i)} / \mu_z , \sigma_z \right) \tag{11}$$

where $P_r$ is the reconstruction probability; $M$ is the number of samples; $p_\theta$ is the data distribution output by the decoder; and $x^{(i)}$ is the $i$-th input sample.

Since the prediction error and the reconstruction probability reflect the degree of prediction deviation and the reconstruction credibility under the normal pattern, respectively, this paper combines the two to define the anomaly score $S_e$:

$$S_e = \frac{e_p}{P_r} \tag{12}$$

To combine the advantages of the prediction network and the reconstruction network, ensure favorable generalization ability of the model, and reduce the potential overfitting risk of the prediction loss and the reconstruction loss, the joint loss function $l_{join}$ of the anomaly detection model is designed as

$$l_{join} = \lambda l_{Trans} + (1 - \lambda) l_{VAE} \tag{13}$$

The specific algorithm pseudocode is shown as follows.

**Algorithm 3** **Transformer-VAE Joint Training and Anomaly Detection**

Input: Training, validation and test windows, feature network parameters, prediction parameters $\theta_p$, VAE parameters $\varphi$, $\theta$, weight $\lambda$

Output: Trained model, detection threshold $\tau$, anomaly score $S_e$ and label $\hat{y}$ of test windows

```
WHILE the training termination criteria are not met DO
    FOR each training batch DO
        Extract spatio-temporal features H according to Algorithm 2;
        Transformer(H;θp) outputs prediction yp;
        VAE encoder outputs μ, σ, sample ε~N(0, I), set z=μ+ε⊙σ;
        Decoder outputs pθ(x|z);
        Set ljoin=λlTrans+(1−λ)lVAE, update parameters for joint training;
    END FOR
END WHILE
Fix model parameters and compute Se;
Determine τ based on POT and fix it for test windows;
FOR each test window DO
    Extract H, compute ep and Pr, set Se=ep/Pr;
    IF Se>τ THEN ŷ←1 ELSE ŷ←0 END IF
END FOR
RETURN trained model, τ, Se, ŷ.
```

### *C. TASHAP: Continuous-interval Temporal Attribution*

The anomaly detection model can output anomaly indicators and identify anomalous windows. However, it is still necessary to further reveal the temporal evolution of faults and quantify the contribution of each time segment and each sensor to anomaly decisions. This paper proposes a continuous time-segment temporal attribution mechanism (TASHAP), which calculates Shapley contributions over the complete fault evolution period to locate high-contribution time segments and critical sensors.

Given an instance window $X$, a set of training background windows $\{X_b\}_{b=1}^{B}$ ($B=2$), and a coalition $S$, the mixing function $H(X, X_b, S)$ retains the instance time segments within $S$, while the remaining time segments are replaced by the same background window.

$$v(S)=\frac{1}{B}\sum_{b=1}^{B} f_{\theta}\left[H(X, X_b, S)\right] \tag{14}$$

where $X$ denotes the instance window to be explained; $\{X_b\}_{b=1}^{B}$ represents the set of training background windows with $B=2$; $S$ is the coalition of time segments; $H(X, X_b, S)$ is the mixing function that preserves the instance time segments in coalition $S$ and substitutes the remaining time segments with background windows; $f_{\theta}$ is the detection model to be interpreted; and $v(S)$ is the value function of coalition $S$, defined as the model output averaged over the $B$ backgrounds under mixed inputs.

After obtaining the values of all coalitions, the Shapley value of time segment $m$ is defined as the weighted average of the marginal contributions generated when this time segment is added to all possible coalitions:

$$\varphi_m=\sum_{S\subseteq N\setminus\{m\}}\frac{|S|!(M-|S|-1)!}{M!}\left[v(S\cup\{m\})-v(S)\right] \tag{15}$$

where $N=\{1,2,\ldots,M\}$ is the set consisting of all time segments; $|S|$ is the number of time segments contained in coalition $S$; $v(S\cup\{m\})-v(S)$ is the marginal contribution of time segment $m$ when it is added to coalition $S$; and $\varphi_m$ is the Shapley value of time segment $m$, which characterizes the average contribution of this time segment to the model's anomaly judgement.

The specific algorithm pseudocode is shown as follows.

| **Algorithm 4** | **TASHAP Continuous Time-Segment and Sensor Attribution** |
|---|---|
| Input | Candidate window X, detection function f, training background $\{X_b\}$ (B=2), M=6, d=48, number of permutations R=4 |
| Output | Time-segment contributions $\{\varphi_m\}$ |
| | Partition X into M continuous segments $P_1,\ldots P_M$ according to equidistant boundaries, and set N={1,…,M}; |
| | **FOR** each coalition S⊆N **DO** |
| | Keep the instance segments within S, and replace the remaining segments with each background $X_b$; |
| | Obtain v(S); |
| | **END FOR** |
| | **FOR** m=1,…,M **DO** |
| | Aggregate the weighted marginal contributions of m joining each coalition to obtain $\varphi_m$; |
| | **END FOR** |
| | **RETURN** $\{\varphi_m\}$. |

### *D. Knowledge-Constrained Auxiliary Root Cause Diagnosis*

The knowledge-constrained auxiliary root cause diagnosis module PKER designed in this paper constrains the judgment and rationale generation of the large language model with a five-dimensional structured evidence package, enabling the traceability and substantiation of traceability conclusions. PKER serves as a flight-level evidence module for the traceability chain. It reads the evidence D1, G1, T1, S1, and C1 generated by anomaly detection and TASHAP, and organizes these structured fields into traceable rationales.

For a candidate window x, TASHAP first generates the evidence package. E(x)={D1(x), G1(x), T1(x), S1(x), C1(x)}. where D1={anomaly indicator, POT threshold, number of consecutive threshold-exceeding windows}; G1={dynamic graph edge density, node summary}; T1={six-segment Shapley values}; S1={top-5 sensor attributions for critical time periods}; C1={command, attitude, velocity, wind estimation summary}. The pseudocode for the algorithm is presented below.

| **Algorithm 5** | **PKER Evidence-Constrained Auxiliary Diagnosis** |
|---|---|
| Input | $D_1$, $G_1$, $T_1$, $S_1$, $C_1$, domain knowledge prompt |
| Output | Diagnosis state $z_{PKER}$, confidence, evidence ID set and diagnosis rationale, response validity |
| 1 | Construct $E=\{D_1,G_1,T_1,S_1,C_1\}$, and associate it with the corresponding voyage using an anonymous case ID; |
| 2 | Write domain knowledge and output constraints into the system prompt, and write E into the user prompt; |
| 3 | Invoke Qwen to obtain decision, confidence, evidence_ids, rationale; |
| 4 | Parse the type, confidence range and evidence ID attribution of the current case; |
| 5 | Judge whether the citations and numerical statements in the diagnosis rationale correspond to E; |
| 6 | **IF** the response satisfies field and evidence constraints **THEN** |
| 7 | Retain decision∈{fault, normal, abstain} together with its confidence, citation set and rationale; |
| 8 | Write the diagnosis result into the corresponding voyage record and mark the response as valid; |
| 9 | **ELSE** |
| 10 | Mark the response as invalid, and keep this status for valid response rate statistics; |
| 11 | **END IF** |
| 12 | **RETURN** diagnosis results and response validity. |

## IV. Experimental results

### *A. Data Introduction and Preprocessing*

The dataset is derived from publicly available real fixed-wing flight records of ALFA, covering telemetry data and fault annotations for key subsystems including propulsion, actuation, navigation control, and environmental perception. Among the 47 flight sessions included in the analysis, 36 are faulty flights, 10 are fault-free flights, and one lacks ground-truth annotations. The 46 labeled flight sessions are adopted for supervised training and evaluation. After alignment at 10 Hz, a total of 46,885 time instants are obtained. The fault state annotations contain 23 engine, 8 aileron, 4 rudder, and 2 elevator labels. A single flight session may have multiple subsystem labels simultaneously, as shown in Fig. 4. The ALFA dataset is publicly available at https://theairlab.org/alfa-dataset/.

After alignment and cleaning, sliding windows are constructed with L=40 and d=48. At a sampling rate of 10 Hz, L=40 corresponds to an observation duration of 4 s, and the interval between window ends is 0.5 s. These windows are used for subsequent window-level anomaly discrimination and continuous temporal attribution. At the feature level, the raw telemetry data have 31 dimensions, mainly including variables such as attitude angles, angular rates, flight velocity, airspeed, control surface deflection angles, engine rotational speed, IMU and navigation states. Owing to asynchronous sampling and data quality discrepancies among different sensors, four asynchronous observation quality metrics, seven physical residual indicators, and six rate-of-change or derived features are further supplemented, yielding 48-dimensional input features.

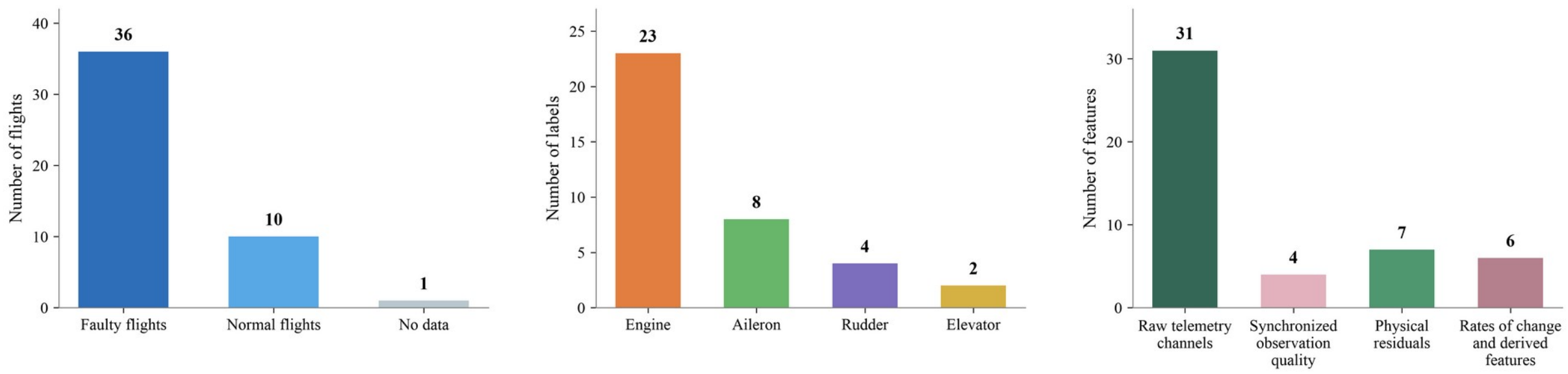

(a) Flight composition (b) Fault-type labels (c) 48-dimensional feature composition

Fig. 4. Data description.

### *B. Evaluation Metrics and Experimental Setup*

Detection evaluation metrics include AUPRC, ACC, Recall and F1. For PKER, fault judgment and evidence citation are evaluated at the flight-session level separately, and these metrics are not mixed with window-level detection metrics. For flight-level diagnosis, PKER calculates Recall and F1 for fault judgment, as well as the G1 graph-edge evidence recall: the proportion of valid G1 evidence IDs cited in the output rationales among valid evidence packages containing G1 summaries.

The experimental environment adopts Python 3.12.7, PyTorch 2.6.0+cu124, scikit-learn 1.5.1 and NumPy 1.26.4. The GPU used is an NVIDIA GeForce RTX 4070 8 GB. In the joint loss function of Transformer-VAE, the weighting coefficient λ is set to 0.7. The learnable weight w_{d} for static-dynamic adjacency matrix fusion is initialized to 0.5 and optimized end-to-end during training. The sliding window size for sensor time series is set to 40 samples with a step size of 5 samples. For the LLM component, we adopt Qwen-plus, with temperature = 0, top-p = 0.7.

### *C. Experimental Results and Analysis*

To verify the effectiveness of the proposed TGTV method, the detection performance of the primary detection model is compared with 11 baselines, and the results are shown in the Table I.

TABLE I

COMPARISON OF DETECTION PERFORMANCE AMONG BASELINE METHODS

| Model | AUPRC | ACC | Recall | F1 |
|---|---|---|---|---|
| LightGBM | 0.972 | 0.964 | 0.805 | 0.880 |
| DAGMM | 0.963 | 0.966 | 0.858 | 0.898 |
| XGBoost | 0.963 | 0.948 | 0.738 | 0.827 |
| RandomForest | 0.956 | 0.947 | 0.721 | 0.822 |
| LSTM | 0.947 | 0.961 | 0.811 | 0.879 |
| GRU | 0.965 | 0.945 | 0.701 | 0.811 |
| GNN | 0.953 | 0.945 | 0.720 | 0.802 |
| OC-CNN | 0.945 | 0.946 | 0.701 | 0.780 |
| DCGL | 0.799 | 0.903 | 0.452 | 0.612 |
| GAN | 0.779 | 0.901 | 0.434 | 0.597 |
| LSTM-VAE | 0.768 | 0.901 | 0.519 | 0.606 |
| TGTV(Ours) | 0.976 | 0.973 | 0.902 | 0.920 |

It can be observed from Table I that AUPRC, ACC, Recall and F1 of the proposed TGTV are 0.976, 0.973, 0.902 and 0.920, respectively. All four metrics are the highest in the table. AUPRC is used to characterize the overall ranking capability of continuous anomaly scores, while Recall and F1 are adopted to reflect the classification performance under the selected threshold. The consistently high values of these metrics indicate that favorable anomaly ranking capability and comprehensive detection performance are achieved by the proposed model. For more intuitive visualization, the comparison between the proposed method and baseline methods is illustrated in the Fig. 5.

It can be seen that DAGMM is the baseline with the highest F1 score in Table I, whose ACC, Recall and F1 are 0.966, 0.858 and 0.898, respectively. To further examine the superiority of the proposed method over other approaches, the relative improvement is illustrated in the Fig. 6.

Compared with DAGMM, the ACC of the proposed TGTV model is increased by 0.7 percentage points, the Recall is increased by 4.4 percentage points, and the F1 is increased by 0.022 (a relative improvement of 2.45%). The missed detection rate is reduced from 14.2% to 9.8%, with a relative reduction of approximately 30.99%. Therefore, the improvement in F1 in this comparison is mainly attributed to more sufficient anomaly detection. When compared with LightGBM, the Recall of the proposed model is raised from 0.805 to 0.902, and the F1 is raised from 0.880 to 0.920, demonstrating superior anomaly detection capability and overall performance. The Recall and F1 of OC-CNN are reported as 0.701 and 0.780, respectively, which are lower than those of the proposed model.

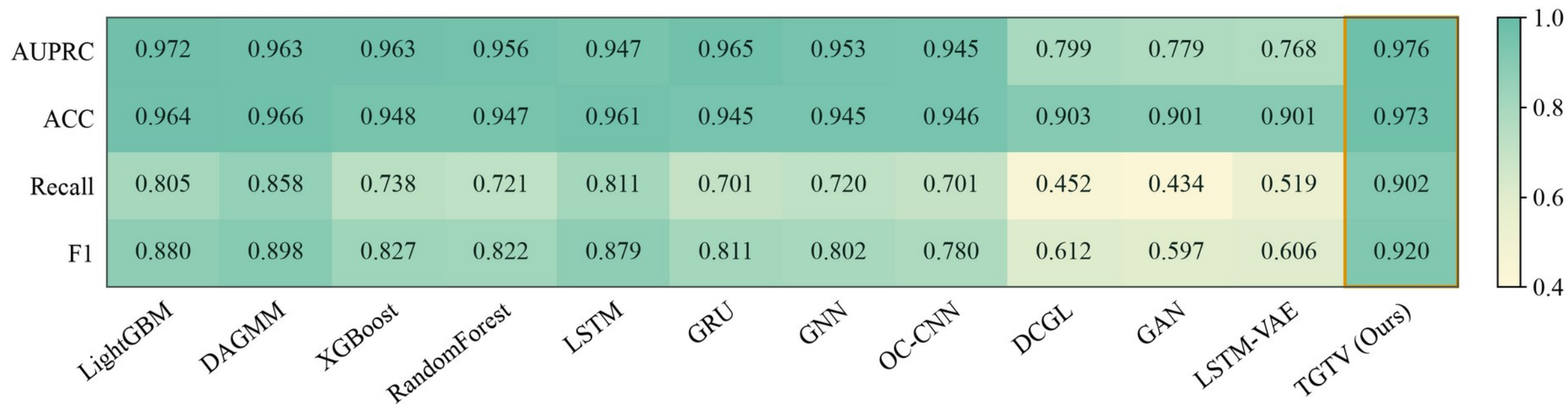

Fig. 5. Comparison between the proposed method and baselines.

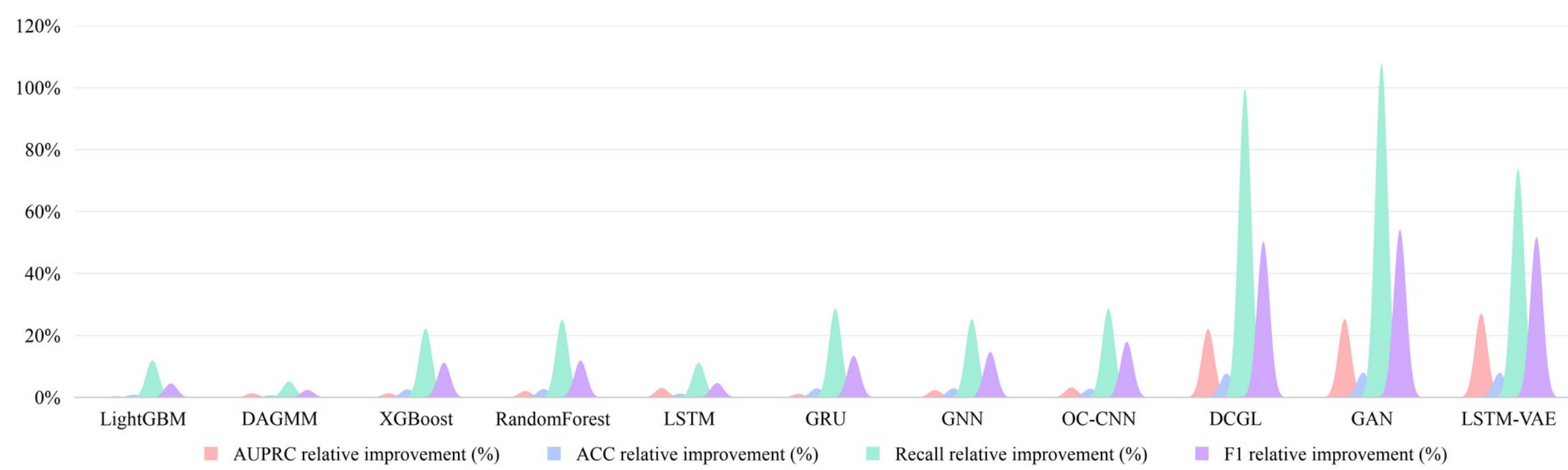

Fig. 6. Relative improvement results.

The Table II presents the complete configuration and the results obtained after removing Transformer, TCN and GAT, respectively.

TABLE II

ABLATION EXPERIMENTS OF COMPONENTS

| Configuration Variant | AUPRC | F1 |
|---|---|---|
| TGTV | 0.976 | 0.920 |
| w/o Transformer | 0.694 | 0.533 |
| w/o TCN | 0.688 | 0.526 |
| w/o GAT | 0.705 | 0.548 |

It can be observed that when the TCN module is removed, the model achieves an AUPRC of 0.688 and an F1 score of 0.526. Compared with the full model, AUPRC suffers an absolute drop of 0.288 and a relative decrease of 29.51%, while F1 has an absolute drop of 0.394 and a relative decrease of 42.83%. The performance degradation is the most significant among the three ablation variants. TCN captures local temporal patterns of telemetry sequences via multi-scale dilated causal convolutions, which can effectively characterize signal fluctuations, slow parameter drifts and persistent shift features during fault evolution. The experimental results demonstrate that local temporal evolution features induced by UAV faults serve as an important foundation for high-precision anomaly discrimination, and multi-scale temporal convolution plays a fundamental role in this spatiotemporal feature extraction framework.

After removing the Transformer module, the model's AUPRC and F1 drop to 0.694 and 0.533, with absolute reductions of 0.282 and 0.387 and relative declines of 28.89% and 42.07%, respectively. Unlike TCN, which focuses on local short-term temporal features, Transformer models long-range temporal dependencies within windows via the multi-head self-attention mechanism to explore intrinsic correlations among flight states at different time steps. Removing either TCN or Transformer leads to considerable performance degradation, indicating that local fine-grained temporal feature extraction and window-level long-range dependency modeling are strongly complementary. Together they enable sufficient mining of information in the temporal dimension.

For the variant without the GAT module, AUPRC and F1 reach 0.705 and 0.548, representing absolute decreases of 0.271 and 0.372 and relative declines of 27.77% and 40.43% compared with the full model. With a static-dynamic weighted fused graph adjacency matrix, the GAT module models dynamic coupling relationships and potential fault propagation among multiple sensors and subsystems. Although the performance drop after removing this module is slightly smaller than that of the two temporal modules, the obvious decrease in metrics verifies that cross-sensor spatial association information can effectively augment temporal features. Meanwhile, the graph relationships output by GAT serve as the source of G1 evidence in the subsequent five-dimensional evidence package. It not only supports the detection task but also provides structured association foundations for fault tracing.

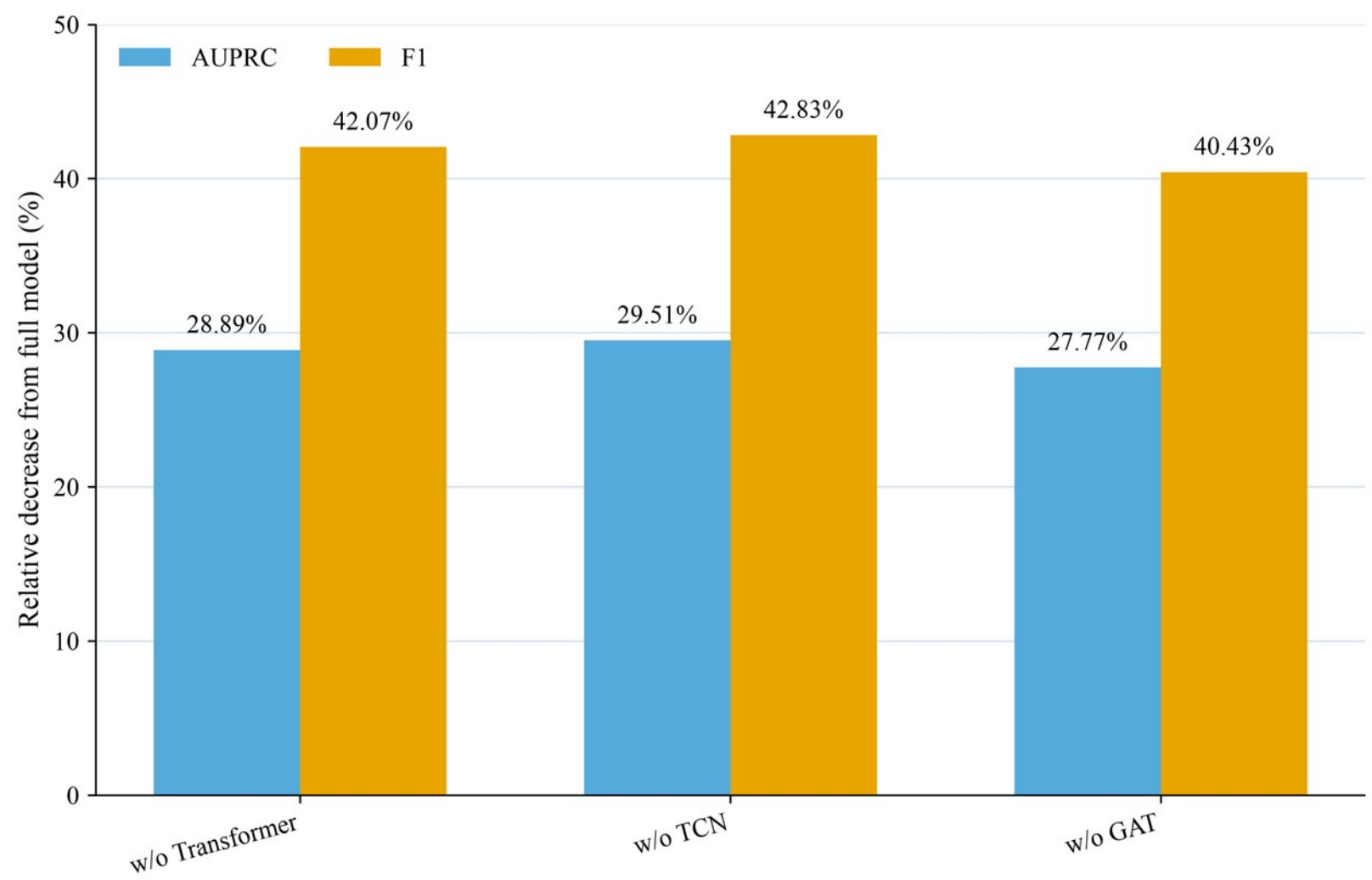

Fig. 7. Relative performance improvements of ablation experiments.

To verify the effectiveness of the proposed PKER method, this paper compares it with the ungrounded large language model (ungrounded LLM) and the numerical rule baseline (without LLM). The evaluation metrics include Recall, F1-score, Traceability and Valid reply rate. The experimental results are shown in Table III.

TABLE III

COMPARATIVE EXPERIMENTS OF LARGE-MODEL TRACEABILITY DIAGNOSIS

| Configuration | Recall | F1 | Traceability | Valid reply |
|---|---|---|---|---|
| Numerical rule baseline (w/o LLM) | 0.902 | 0.910 | 0.615 | 0.873 |
| Ungrounded LLM | 0.884 | 0.794 | 0.620 | 0.782 |
| Full PKER-LLM (Proposed) | 0.943 | 0.944 | 0.918 | 0.976 |

It can be observed that significant degradation in overall performance is exhibited by the LLM without evidence constraints under unconstrained evidence conditions. Its F1 score and valid response rate are even lower than those of the numerical rule baseline, which indicates that hallucinations are prone to being generated by the large model and insufficient reliability is obtained for its outputs. To further compare the relative improvement, the results are illustrated in the Fig. 8.

Compared with the unconstrained LLM without evidence constraints, positive improvements in all metrics are achieved by the complete PKER model. The recall is relatively improved by 6.67%, the F1 score by 18.89%, and the valid response rate by 24.81%. The most prominent improvement is observed for the traceability metric, with a relative increase of 48.06%. The relatively limited gain in recall indicates that a certain capability of relevant information retrieval is already possessed by the unconstrained LLM. In contrast, the substantial growth in traceability demonstrates that the evidence constraint mechanism introduced in PKER can effectively standardize model outputs, significantly strengthen the correlation between answers and reference evidence, and mitigate model hallucinations. Meanwhile, the notable gains of the PKER model in F1 and valid response rates verify that the evidence constraint strategy can enhance the accuracy and output compliance of model responses. Overall, the PKER model outperforms both the unconstrained LLM and the traditional numerical rule baseline across all evaluation metrics, which validates the superiority of the proposed method. As a supplementary evaluation of diagnostic text quality, 90 diagnostic results (30 samples for each configuration) randomly sampled in a flight-session-stratified manner are independently scored on a five-point scale by three raters with backgrounds in UAV flight control and fault diagnosis.

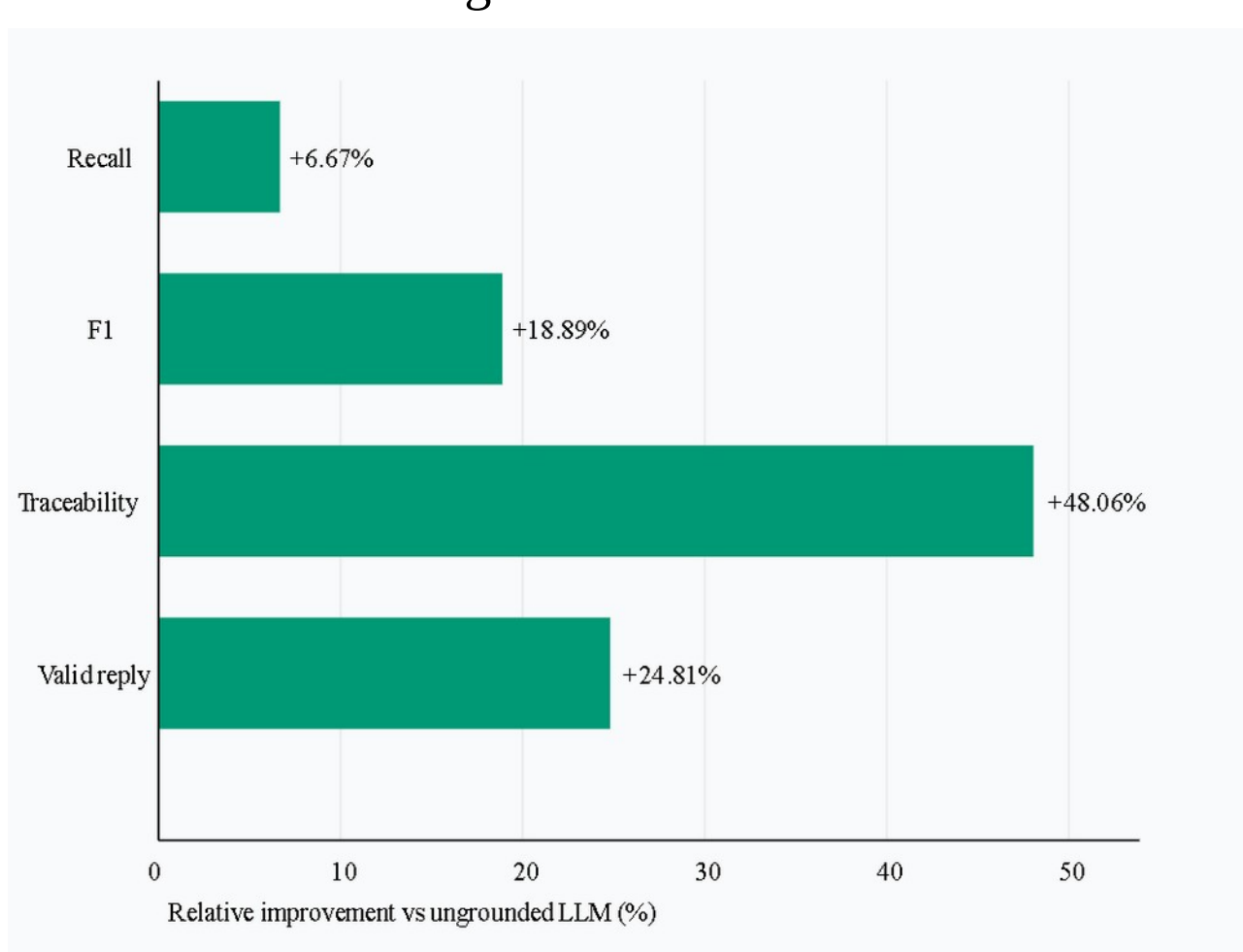

Fig. 8. Relative improvements over the baseline language model.

The metrics of Evidence Support E, Attribution Logic Consistency C, Readability R and Decision Usability U are presented separately in Table IV.

Table IV shows that the ungrounded LLM achieves a readability score of 3.67, which is higher than the 1.33 of the numerical rule method. However, its evidence support, logic consistency and decision usability are only 2.33, 2.00 and 2.17, respectively. This indicates that fluent expression and reliable explanation are two distinct requirements. PKER obtains four scores of 4.67, 4.50, 4.67 and 4.33. Compared with the ungrounded LLM, the scores increase by 2.34, 2.50, 1.00 and 2.16 points. The performance gains are mainly concentrated in evidence support, logic consistency and decision usability. These dimensional results complement the high traceability of evidence IDs in Table III and confirm the importance of evidence organization and citation correspondence in diagnostic texts.

TABLE IV

EXPLAINABILITY SCORES OF LARGE-MODEL TRACEABILITY

| Evaluation Dimension | Numerical rule (w/o LLM) | Ungrounded LLM | Full PKER-LLM |
|---|---|---|---|
| Evidence Support E | 1.67 | 2.33 | 4.67 |
| Attribution Logic Consistency C | 1.83 | 2.00 | 4.50 |
| Readability R | 1.33 | 3.67 | 4.67 |
| Decision Usability U | 2.00 | 2.17 | 4.33 |

## V. Discussion and Conclusion

### *A. Discussion*

The proposed model achieves the highest AUPRC, ACC, Recall and F1, demonstrating superior anomaly detection performance. Compared with DAGMM, the baseline with the best F1 score, our method improves Recall by 4.4 percentage points, reduces the window-level missing detection rate from 14.2% to 9.8%, and increases F1 by 2.2 percentage points. This result shows that the model achieves a favorable balance between anomaly detection capability and overall performance, and reflects the complementary effects of spatiotemporal representation and hybrid detection. The value of PKER lies in organizing numerical detection results and attribution summaries into evidence-based voyage diagnostic records. Compared with the ungrounded LLM, PKER improves F1 by 15 percentage points, and the dimensional scores also reveal its stronger evidence support and logic consistency.

The framework provides model-level explanations and associative traceability clues. The edge weights of GAT reflect variable dependencies learned by the model, and TASHAP characterizes output contributions relative to the training background. These results offer associative cues for fault localization. Valid evidence IDs maintain the correspondence between diagnostic reasoning and input fields, whereas graph edges and Shapley contributions remain model-level associative information. Future research may start with offline fault analysis and auxiliary diagnosis, and then advance online deployment with more extensive datasets and operational tests.

### *B. Conclusion*

Aiming at the requirements of fault detection and explanation in UAV flight telemetry, this paper proposes TGTV, a large-model-based interpretable UAV fault detection and traceability framework integrating spatiotemporal graph learning. The detection pipeline outputs window-level anomaly results, while the attribution and auxiliary diagnosis pipeline further organizes evidence of time segments, sensors, graph relationships and working conditions to provide structured records for fault analysis. In the ALFA cross-voyage comparison experiments, TGTV achieves AUPRC, ACC, Recall and F1 of 0.976, 0.973, 0.902 and 0.920, respectively. Compared with DAGMM, the baseline with the highest F1, TGTV improves Recall by 4.4 percentage points and F1 by 0.022, demonstrating favorable anomaly detection and overall performance. For voyage diagnosis, PKER obtains a Recall of 0.943, an F1 of 0.944, an evidence ID traceability rate of 0.918 and a valid reply rate of 0.976. The dimensional evaluation further verifies that its explanations possess stronger evidence support, logic consistency and decision usability.

Overall, the results in this paper demonstrate that anomaly detection, sequential attribution and structured linguistic diagnosis can work collaboratively within a unified framework to detect and analyze anomalies. Graph relationships and attribution results provide valuable associative clues for fault localization. Combined with maintenance records and flight control mechanisms, root cause identification and online applications can be further promoted.

## REFERENCES

[1] Y. Cui, Y. Liang, J. Shi, C. Ma, H. Zhang, and S. Li, "An interpretable hierarchical framework for rapid UAV-type recognition based on multi-sensor data and knowledge," *IEEE Sensors J.*, vol. 25, no. 12, pp. 22239–22249, 2025, doi: 10.1109/jsen.2025.3565307.

[2] M. W. Ahmad, M. U. Akram, R. Ahmad, K. Hameed, and A. Hassan, "Intelligent framework for automated failure prediction, detection, and classification of mission critical autonomous flights," *ISA Trans.*, vol. 129, pp. 355–371, 2022, doi: 10.1016/j.isatra.2022.01.014.

[3] Z. Cai, Y. Zhang, and Q. Miao, "PCAP-Net: Predictive consistency-aware perception network for anomaly detection in UAV sensor data," *IEEE Sensors J.*, vol. 26, no. 10, pp. 16017–16026, 2026, doi: 10.1109/jsen.2026.3682503.

[4] A. Keipour, M. Mousaei, and S. Scherer, "ALFA: A dataset for UAV fault and anomaly detection," *Int. J. Robot. Res.*, vol. 40, no. 2–3, pp. 515–520, 2021, doi: 10.1177/0278364920966642.

[5] J. D. Yoo, G. M. Kim, M. G. Song, and H. K. Kim, "MeNU: Memorizing normality for UAV anomaly detection with a few sensor values," *Comput. Secur.*, vol. 150, Art. no. 104248, 2025, doi: 10.1016/j.cose.2024.104248.

[6] J. Wang, Q. Liao, Z. Gu, and L. Nie, "Anomaly detection of UAV flight data based on spatiotemporal feature fusion," *Appl. Soft Comput.*, vol. 197, Art. no. 115131, 2026, doi: 10.1016/j.asoc.2026.115131.

[7] Z. Gu, J. Wang, J. Wang, and L. Nie, "Spatial-temporal correlation-based anomaly detection in UAV flight data" (in Chinese), *J. Xidian Univ.*, vol. 52, no. 4, pp. 33–45, 2025, doi: 10.19665/j.issn1001-2400.20250403.

[8] J. Liu, J. Fan, and G. Jin, "A novel method for generating small-sample data of the satellite power subsystem with InCoTimeGAN integrating self-checking and information enhancement," *IEEE Sensors J.*, vol. 26, no. 17, pp. 26081–26091, 2026, doi: 10.1109/jsen.2026.3707885.

[9] T. Yang, Y. Lu, H. Deng, and C. Tang, "A hybrid multimodal neural-network-based anomaly detection model for UAVs," *IEEE Trans. Aerosp. Electron. Syst.*, vol. 61, no. 4, pp. 10273–10291, 2025, doi: 10.1109/taes.2025.3562543.

[10] Y. He, Z. Lu, Y. Deng, and D. Wang, "FG-DeformT: Grouped deformable temporal modeling for flight-level aircraft anomaly detection," *Sensors*, vol. 26, no. 10, Art. no. 3189, 2026, doi: 10.3390/s26103189.

[11] C. Feng, J. Fan, Z. Liu, G. Jin, and S. Chen, "Unmanned aerial vehicle anomaly detection based on causality-enhanced graph neural networks," *Drones*, vol. 9, no. 6, Art. no. 408, 2025, doi: 10.3390/drones9060408.

[12] F. Hu, Q. Wang, H. Shao, S. Gao, and H. Yu, "Anomaly detection of UAV state data based on single-class triangular global alignment kernel extreme learning machine," *Comput. Model. Eng. Sci.*, vol. 136, no. 3, pp. 2405–2424, 2023, doi: 10.32604/cmes.2023.026732.

[13] X. Liu *et al.*, "UTIDH: Unstable Transformer with imbalanced data handling for UAV multivariate time series fault diagnosis," *IEEE Trans. Veh. Technol.*, vol. 75, no. 6, pp. 11199–11208, 2026, doi: 10.1109/tvt.2025.3636096.

[14] J. Chen, D. Zhang, Z. Jin, L. Qiao, and H. Zhang, "UAV-EGAT: An enhanced graph attention network for UAV elevon fault detection," *Syst. Soft Comput.*, vol. 8, Art. no. 200481, 2026, doi: 10.1016/j.sasc.2026.200481.

[15] S. Lyu, K. Wang, Y. Liu, and B. Wang, "Hierarchical GNN message passing for node-level anomaly detection in industrial control systems," *IEEE Trans. Dependable Secure Comput.*, vol. 23, no. 2, pp. 1994–2011, 2026, doi: 10.1109/tdsc.2025.3623299.

[16] R. Dhakal, C. Bosma, P. Chaudhary, and L. N. Kandel, "UAV fault and anomaly detection using autoencoders," in *Proc. IEEE/AIAA 42nd Digit. Avionics Syst. Conf. (DASC)*, Oct. 2023, pp. 1–8, doi: 10.1109/dasc58513.2023.10311126.

[17] C. Feng, J. Fan, G. Jin, J. Liu, and S. Chen, "STE-RS: A novel framework for robust anomaly detection in UAV flight data," *Results Eng.*, vol. 27, Sep. 2025, Art. no. 106801, doi: 10.1016/j.rineng.2025.106801.

[18] X. Chen, H. Xiao, Z. Zeng, S. Zhang, and R. Xiao, "Fine-grained multivariate time series anomaly detection via causal inference," *Knowl.-Based Syst.*, vol. 331, Jan. 2026, Art. no. 114765, doi: 10.1016/j.knosys.2025.114765.

[19] C. Jiang and W. Wu, "Anomaly detection in distributed systems based on spatio-temporal causal inference," *Eng. Appl. Artif. Intell.*, vol. 160, pt. A, Nov. 2025, Art. no. 111857, doi: 10.1016/j.engappai.2025.111857.

[20] X. Chen, R. Xiao, Z. Zeng, S. Zhang, and X. Du, "Semi-supervised anomaly detection via reinforcement learning-enabled method with causal inference," *Inf. Sci.*, vol. 719, Nov. 2025, Art. no. 122463, doi: 10.1016/j.ins.2025.122463.

[21] S. Xing, Y. Wang, and W. Liu, "Multi-dimensional anomaly detection and fault localization in microservice architectures: A dual-channel deep learning approach with causal inference for intelligent sensing," *Sensors*, vol. 25, no. 11, May 2025, Art. no. 3396, doi: 10.3390/s25113396.

[22] H. Ali, J. Liu, and F. Gao, "A hybrid causal-inference and neuro-fuzzy framework for advanced process monitoring," *J. Process Control*, vol. 162, Jun. 2026, Art. no. 103724, doi: 10.1016/j.jprocont.2026.103724.

[23] Z. Zeng, X. Deng, and B. Zhao, "Causal gray wolf optimization: A novel approach to robust network anomaly detection amidst data redundancy," *IEEE Internet Things J.*, vol. 12, no. 24, pp. 52820–52836, Dec. 2025, doi: 10.1109/jiot.2025.3613994.

[24] J. Liu, G. Jin, S. Chen, J. Fan, and Y. Han, "Knowledge data fusion via LLM for domain-specific small-sample causal discovery," *Knowl.-Based Syst.*, vol. 351, Oct. 2026, Art. no. 116636, doi: 10.1016/j.knosys.2026.116636.

[25] Z. Zhao, H. Yang, Y. Zhu, and Q. Feng, "IMDM-PC: Incremental multi-scale decomposition–mixing with physics consistency for LLM-based ADS-B trajectory anomaly detection," *Comput. Netw.*, vol. 286, Aug. 2026, Art. no. 112476, doi: 10.1016/j.comnet.2026.112476.

[26] X. Jiang, C. Shen, B. Yao, and S. Shao, "GCRdetector: Graph contrastive representation learning for urban anomaly detection," *Inf. Sci.*, vol. 748, Aug. 2026, Art. no. 123490, doi: 10.1016/j.ins.2026.123490.

[27] Z. Zhang, S. Li, L. Zhang, J. Ye, C. Hu, and L. Yan, "LLM-LADE: Large language model-based log anomaly detection with explanation," *Knowl.-Based Syst.*, vol. 326, Sep. 2025, Art. no. 114064, doi: 10.1016/j.knosys.2025.114064.

[28] J. L. T. Lau *et al.*, "BiTAD: An interpretable temporal anomaly detector for 5G networks with TwinLens explainability," *Future Internet*, vol. 17, no. 11, Oct. 2025, Art. no. 482, doi: 10.3390/fi17110482.

[29] V. V. M. Anees, K. P. A. Nazar, R. Sridharan, and S. Maniyath, "Explainable and secure wind power forecasting: SHAP-enhanced framework with feature stability, anomaly detection, and adversarial robustness," *Appl. Soft Comput.*, vol. 196, Jun. 2026, Art. no. 115091, doi: 10.1016/j.asoc.2026.115091.

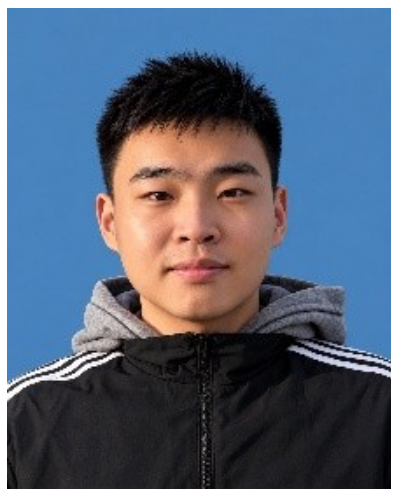

**ZIXUAN WANG** is currently pursuing a master's degree in data science with the School of Computer Science, Faculty of Engineering, The University of Sydney, Sydney, NSW, Australia. His research interests include machine learning, data analytics, and artificial intelligence, with a focus on data-driven fault detection and interpretable analysis of complex systems.

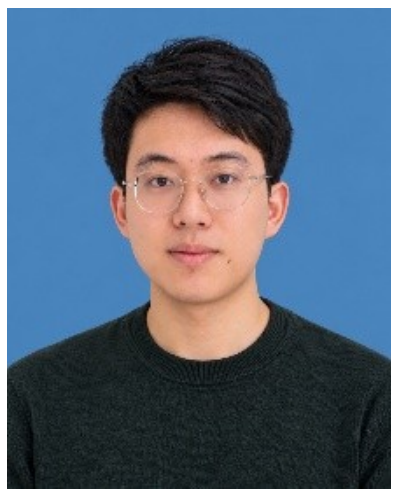

**CHANGBO SHI** is currently pursuing a master's degree in computer science with the School of Computer Science, Faculty of Engineering, The University of Sydney, Sydney, NSW, Australia. His research interests include embodied artificial intelligence and learning-based methods for intelligent systems.

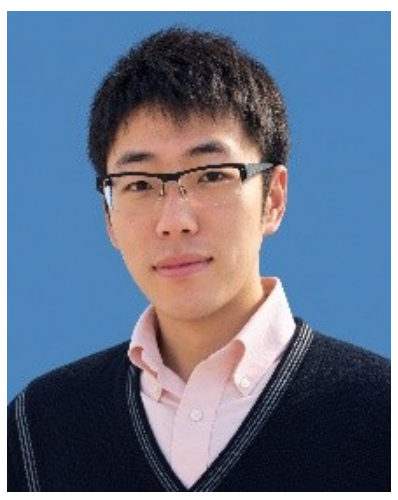

**JI LIU** received the M.Com. and Ph.D. degrees from The University of Sydney, Sydney, NSW, Australia. He is currently a Postdoctoral Research Associate with the School of Electrical and Computer Engineering, Faculty of Engineering, The University of Sydney. His research interests include artificial intelligence, distributed machine learning, blockchain, and financial technology. His work applies machine learning and data analytics to prediction and risk monitoring, including distributed learning for medical data analysis, financial time-series forecasting, and blockchain-based financial systems.